\documentclass{article}
\usepackage[utf8]{inputenc}
\usepackage{amsmath, amssymb}
\usepackage{graphicx}
\usepackage{booktabs}
\usepackage{hyperref}
\usepackage{geometry}
\usepackage{natbib}
\usepackage{placeins}
\usepackage{float}
\usepackage{algorithm}
\usepackage{algpseudocode}

\graphicspath{{figures/}}
\title{Modeling and Controlling Deployment Reliability under Temporal Distribution Shift}

\author{
Naimur Rahman \\
\texttt{n.rahman@bathspa.ac.uk}
\and
Naazreen Tabassum \\
\texttt{naazreentabassum12nn@gmail.com}
}

\date{1/3/2026}

\begin{document}

\maketitle

\begin{abstract}
Machine learning systems deployed in non-stationary environments experience a temporal distribution shift that alters the statistical relationship between inputs and outcomes. Although existing mitigation strategies such as retraining, recalibration, and drift detection aim to preserve predictive performance, they typically evaluate models at isolated time points and optimize average metrics. This neglects the temporal stability of reliability during deployment.

We introduce a formal deployment framework in which the reliability is modeled as a dynamic state comprising discrimination, calibration, and stability components. The evolution of this state over time windows induces a trajectory whose volatility can be explicitly quantified. We cast deployment adaptation as a multi-objective control problem that minimizes reliability volatility subject to intervention cost constraints.

Within this framework, we define a class of intervention policies and construct the empirical Pareto frontier in the cost-volatility space. Experiments on a temporally indexed credit-risk dataset demonstrate that selective, state-dependent interventions can achieve lower reliability volatility than continuous retraining while substantially reducing operational cost.

These findings establish deployment reliability under temporal shift as a controllable multi-objective system and provide a principled basis for designing intervention policies in high-stakes tabular applications.
\end{abstract}

\section{Introduction}

Machine learning systems deployed in operational environments are routinely exposed to temporal distribution shift. In high-stakes tabular domains such as credit risk assessment, fraud detection, and healthcare decision support, data-generating processes evolve due to economic cycles, regulatory intervention, demographic change, and feedback effects induced by model-driven decisions. As a result, models validated under historical data may degrade or behave unpredictably when evaluated on temporally forward cohorts. This raises the question: how should we design deployment policies that explicitly control reliability over time under temporal distribution shift?

A substantial literature studies dataset shift and concept drift. Which proposes mitigation strategies such as periodic retraining, rolling-window updates, recalibration, and statistical drift detection. These approaches are typically evaluated using average predictive metrics—such as area under the ROC curve (AUC) or expected calibration error (ECE)—computed independently at each evaluation window. While effective for diagnosing degradation, they do not explicitly model how predictive reliability evolves across deployment horizons, nor do they formalize the trade-off between intervention frequency and operational cost.

In practice, instability of predictive reliability can be as consequential as mean performance degradation. Abrupt fluctuations in discrimination or calibration complicate governance, increase risk exposure, and undermine trust in automated systems. Monitoring dashboards and drift detectors often trigger interventions heuristically once thresholds are exceeded, but such triggers are rarely grounded in explicit stability objectives or cost-aware optimization criteria. The temporal stability of reliability remains largely unmodeled in a way that jointly accounts for intervention cost.

This paper introduces a deployment-centric framework in which reliability is modeled as a dynamic state indexed by time. The reliability state comprises discrimination and calibration components, and its evolution over deployment windows defines a trajectory whose volatility can be explicitly quantified. We formulate deployment adaptation as a multi-objective control problem: minimize reliability volatility while constraining cumulative intervention cost.

Within this framework, intervention policies operate over the observed reliability state and drift signals. We analyze policies ranging from static deployment and periodic retraining to drift-triggered reliability control (DTRC) and a multi-objective reliability control (MORC) sweep that explores a family of threshold-based policies in cost–volatility space. Experiments on a large-scale, temporally indexed credit-risk dataset (1.35M loans, 2007–2018, evaluated in annual windows) show that selective, state-dependent intervention can dominate frequency-based retraining strategies, achieving lower reliability volatility at substantially reduced operational cost.

\textbf{Contributions.}
\begin{enumerate}
    \item We formalize deployment reliability under temporal distribution shift as a dynamic state, separating average predictive performance from temporal stability.
    \item We introduce volatility-based stability objectives and cast deployment adaptation as a multi-objective control problem under cumulative intervention cost constraints.
    \item We define a class of drift-triggered intervention policies operating over the reliability state and drift indicators, including a concrete Drift-Triggered Reliability Control (DTRC) policy.
    \item We empirically characterize the cost–volatility Pareto frontier induced by deployment policies and identify knee configurations that dominate continuous rolling retraining in stability–cost space, reducing retraining-related cost by approximately 73\% with only modest loss in discrimination.
\end{enumerate}

\section{Related Work}

\subsection{Temporal Dataset Shift and Concept Drift}

Distribution shift arises when the joint distribution of inputs and labels at deployment differs from that observed during training. A comprehensive treatment of dataset shift is provided by \citet{QuioneroCandela2009}, who formalize covariate shift, label shift, and concept shift as distinct mechanisms. Surveys on concept drift further examine adaptation strategies in non-stationary environments \citep{Gama2014, Widmer1996}. Empirical evidence from real-world benchmarks such as WILDS demonstrates that models validated under random splits can degrade substantially under naturally occurring temporal or distributional shifts \citep{Koh2021}.

In tabular risk modeling, temporal variation is structurally expected due to economic cycles and policy interventions. Studies in credit scoring highlight the sensitivity of predictive models to evaluation protocols and temporal partitioning \citep{Lessmann2015, Hand1997}. These works analyze performance degradation under shift but do not model the temporal evolution of reliability metrics as a controlled state variable or optimize the stability of reliability trajectories.

\subsection{Mitigation Strategies under Shift}

Common mitigation strategies include periodic retraining, rolling-window updates, and online learning approaches. Rolling retraining emphasizes recent data to improve adaptability under drift \citep{Gama2014}. Online learning provides regret guarantees in sequential settings but does not typically model intervention cost explicitly \citep{CesaBianchi2006}.

Recalibration methods aim to correct probability estimates without modifying the underlying predictor. Classical approaches include Platt scaling \citep{Platt1999} and isotonic regression \citep{Zadrozny2002}, while temperature scaling has become standard for modern classifiers \citep{Guo2017}. Under label shift, correction procedures adjust predicted probabilities using estimated prior changes \citep{Saerens2002, Lipton2018}. These methods improve pointwise calibration but do not explicitly address how calibration and discrimination co-evolve over time under temporal shift, or how to control the volatility of this reliability trajectory.

Recent work on cost-aware retraining formalizes the binary retrain-or-keep decision as a cost--accuracy trade-off in streaming settings \citep{Mahadevan2024}. However, this approach optimizes a scalar staleness cost rather than modeling reliability as a multi-dimensional state with trajectory stability as an explicit objective. In contrast, we treat discrimination and calibration as components of a time-indexed reliability state and study how different intervention policies shape the volatility of this state across deployment.

\subsection{Drift Detection and Monitoring}

Statistical drift detection methods identify distributional change in streaming data. The Drift Detection Method (DDM) \citep{Gama2004}, Early Drift Detection Method (EDDM) \citep{BaenaGarcia2006}, and ADWIN \citep{Bifet2007} provide sequential hypothesis-testing mechanisms for change detection. Kernel-based two-sample tests such as Maximum Mean Discrepancy offer nonparametric tools for detecting distributional differences \citep{Gretton2012}.

In production machine learning systems, monitoring frameworks track co variate drift, calibration error, and performance degradation over time \citep{Sculley2015, Ovadia2019}. However, these approaches do not model the joint trade-off between stability of reliability trajectories and cumulative intervention cost, nor do they characterize the Pareto frontier over deployment policies.

\subsection{Multi-Objective Control for Deployment Reliability}

Multi-objective optimization formalizes trade-offs among competing criteria through Pareto optimality \citep{Deb2002, Boyd2004}. In machine learning, such approaches have been applied to balance fairness and accuracy or optimize non-decomposable objectives \citep{Eban2017, Zafar2017}. Control theory and reinforcement learning provide frameworks for sequential decision-making under dynamic conditions \citep{Astrom1995, Sutton2018}.

To our knowledge, no prior work models predictive reliability under temporal dataset shift as a dynamic state with volatility as an explicit optimization objective, nor applies multi-objective control to select deployment policies along an empirical cost–volatility Pareto frontier.

\section{Problem Formulation}
\label{sec:problem_formulation}

\subsection{Deployment Setting}

Consider a supervised binary classification problem with input space $\mathcal{X}$ and label space $\mathcal{Y} = \{0, 1\}$. A predictive model $f_t : \mathcal{X} \rightarrow [0, 1]$ produces probabilistic outputs at discrete time indices $t = 1, \ldots, T$. Data arrive in non-overlapping, chronologically ordered evaluation windows corresponding to annual cohorts (here, $T = 9$ years from 2010 to 2018). Let $\mathcal{D}_t \sim P_t(X, Y)$ denote the joint distribution at time $t$. Temporal distribution shift is defined by $P_t \neq P_{t'}$ for some $t \neq t'$. No assumption is made regarding the mechanism of shift; covariate shift, label shift, and concept drift may co-occur.

Actions that include retraining update the underlying model parameters defining $f_t$, while recalibration updates only a calibration map applied to the outputs of $f_t$ without modifying its ranking structure.

\subsection{Reliability as a Dynamic State}

At each time $t$, model performance is evaluated using:
\begin{itemize}
  \item Discrimination $A_t$ measured by area under the ROC curve (AUC),
  \item Calibration error $C_t$ measured by expected calibration error (ECE).
\end{itemize}
We define the reliability state at time $t$ as
\[
R_t = (A_t, C_t).
\]

To capture temporal stability, we summarize reliability over the deployment horizon as the mean absolute change in the reliability state between consecutive evaluation windows. The $L_1$ volatility of reliability is
\[
V_{L1} = \frac{1}{T-1} \sum_{t=2}^{T} \left( |A_t - A_{t-1}| + |C_t - C_{t-1}| \right).
\]
Downside volatility restricts this sum to steps where reliability degrades, e.g.\ where $A_t < A_{t-1}$ or $C_t > C_{t-1}$:
\[
V^-_{L1} = \frac{1}{T-1} \sum_{t=2}^{T} \mathbf{1}\{\Delta_t < 0\} \left( |A_t - A_{t-1}| + |C_t - C_{t-1}| \right),
\]
where $\Delta_t$ denotes a chosen scalar summary of the reliability change (for example, $\Delta_t = A_t - A_{t-1}$ or a weighted combination of changes in $A_t$ and $C_t$).

This formulation treats reliability as a trajectory rather than a set of isolated pointwise evaluations, and volatility as the temporal instability of that trajectory.

\subsection{Intervention Policies and Cost}
\label{sec:problem_cost}

At each time $t$, a policy $\pi$ selects an action from
\[
\mathcal{A} = \{\text{NoOp}, \text{Recalibrate}, \text{Retrain}, \text{Both}\}.
\]
We define deployment intervention cost to reflect the dominant operational burden of retraining (compute, validation, and governance) relative to recalibration. The per-window cost is
\[
c(a_t) =
\begin{cases}
5, & \text{if } a_t = \text{Retrain},\\
6, & \text{if } a_t = \text{Both},\\
1, & \text{if } a_t = \text{Recalibrate},\\
0, & \text{if } a_t = \text{NoOp}.
\end{cases}
\]
Total deployment cost under policy $\pi$ is
\[
C(\pi) = \sum_{t=1}^{T} c(a_t),
\]
which measures the cumulative weighted intervention cost over the deployment horizon.

\subsection{Multi-Objective Deployment Problem}

Deployment adaptation is formulated as a multi-objective optimization problem:
\[
\min_{\pi} \big( V_{L1}(\pi),\, C(\pi) \big).
\]
Rather than optimizing a single scalar objective, we characterize the Pareto frontier in cost–volatility space and analyze policies in terms of dominance and trade-off structure.

\section{Method}
\label{sec:method}

\subsection{From Temporal Evaluation to Deployment Policies}

We consider deployment as a sequential decision process in which the system is evaluated in discrete, non-overlapping annual windows and may be updated between windows. Let $R_t = (A_t, C_t)$ denote the observed reliability state at window $t$, where $A_t$ is discrimination (ROC AUC) and $C_t$ is calibration error (ECE). A deployment policy $\pi$ maps the observed reliability history and drift signals to an intervention action:
\[
\pi: \{R_1, \dots, R_t, D_1, \dots, D_t\} \rightarrow \mathcal{A},
\quad \mathcal{A} = \{\text{NoOp}, \text{Recalibrate}, \text{Retrain}, \text{Both}\}.
\]
Reliability metrics $(A_t, C_t)$ are computed from labeled outcomes in window $t$ after outcomes are realized, and any intervention selected at time $t$ updates the deployed system for subsequent windows ($t+1$ onward). This protocol avoids look-ahead: decisions at time $t$ use only information available up to $t$.

\paragraph{Action semantics.}
Each action corresponds to an operational update with distinct scope:
(i) \textbf{NoOp} keeps the predictor unchanged;
(ii) \textbf{Recalibrate} fits a post-hoc calibration map (isotonic regression) to model scores using labeled data from the previous window $t{-}1$, preserving ranking while adjusting probability estimates;
(iii) \textbf{Retrain} refits the underlying predictor on a rolling historical window of length $W$ years ($W{=}3$ in our experiments), potentially altering both ranking and probabilities; and
(iv) \textbf{Both} retrains and then recalibrates the updated predictor.
We use the weighted action-cost model defined in Section~\ref{sec:problem_cost} to reflect operational burden.

\paragraph{Cost model abstraction.}
We model operational cost through a weighted scheme that charges retraining at five times the cost of recalibration, reflecting the dominant engineering, compute, and governance burden of model refitting relative to post-hoc calibration adjustment. In practice, costs may also depend on compute budgets, validation workflows, regulatory oversight, and model risk management procedures. Extending the framework to richer, domain-specific cost models is an important direction for deployment realism.

\subsection{Baseline Policies}
\label{sec:baselines}

We evaluate three standard deployment baselines that represent common production heuristics and provide reference points for the stability--cost trade-off.

\paragraph{P0: Static deployment.}
The model is trained once on all data prior to the deployment horizon (here, all loans issued before 2010) and then deployed without further updates ($a_t = \text{NoOp}$ for all $t$). This baseline isolates the effect of temporal distribution shift on reliability trajectories in the absence of any corrective action. Under the weighted cost model, P0 incurs zero deployment cost.

\paragraph{P1: Periodic recalibration.}
The model is trained once before the deployment horizon. At each subsequent window $t$, an isotonic regression calibration map is fitted using labeled data from the previous window $t{-}1$ and applied prospectively to window $t$ ($a_t = \text{Recalibrate}$). Isotonic regression is a non-parametric calibration method that can correct arbitrary monotone distortions in predicted probabilities without changing the model's ranking structure. In high-stakes tabular settings where score distributions may evolve nonlinearly, this flexibility is often preferable to parametric scaling rules such as Platt scaling or temperature scaling.

\paragraph{P2: Rolling retraining.}
At each window $t$, the underlying predictor is retrained using a rolling historical window of length $W{=}3$ years (i.e., training on years $t{-}3$ through $t{-}1$ when available; $a_t = \text{Retrain}$). Rolling windows emphasize recent data and are widely used under drift to avoid overweighting obsolete regimes. We treat rolling retraining as a high-intervention reference policy: it is expected to improve average discrimination under shift, but at substantial operational cost ($C(\pi) = 5 \times T = 45$ over the full deployment horizon).

\begin{table}[H]
\centering
\caption{Deployment policy family evaluated in this study.}
\label{tab:policy_overview}
\begin{tabular}{lll}
\toprule
Policy & Trigger & Update Action \\
\midrule
P0 (Static) & None & NoOp \\
P1 (Periodic recalibration) & Every window & Recalibrate (isotonic, using $t{-}1$ data) \\
P2 (Rolling retraining) & Every window & Retrain (rolling $W{=}3$ years) \\
P3 (DTRC) & Drift \& reliability thresholds & State-dependent (Retrain / Recalibrate / Both) \\
P4 (MORC) & Threshold sweep & Pareto-based policy selection \\
\bottomrule
\end{tabular}
\end{table}

\subsection{Drift-Triggered Reliability Control (DTRC)}
\label{sec:dtrc}

The baseline policies above intervene either never (P0) or according to a fixed schedule (P1--P2). In contrast, DTRC intervenes selectively based on observed evidence of distributional shift and reliability degradation. The key intuition is that not all drift warrants retraining: moderate shifts may primarily distort probability calibration (addressable by recalibration), while severe shifts may compromise ranking itself (requiring retraining). DTRC therefore gates interventions hierarchically using (i)~drift magnitude and (ii)~reliability failure signals.

\paragraph{Drift indicators.}
We summarize distributional change between adjacent windows using separate indicators for numeric and categorical features. For numeric features, we compute the Kolmogorov--Smirnov (KS) two-sample statistic per feature and report the mean across features, $\overline{\mathrm{KS}}_t$. For categorical features, we compute the Jensen--Shannon divergence (JSD) per feature using compressed category histograms (retaining the top-$k$ categories with $k{=}50$ and grouping the remainder into an \texttt{OTHER} bin) and report the mean, $\overline{\mathrm{JSD}}_t$. The combined drift signal is
\[
D_t = \alpha \,\overline{\mathrm{KS}}_t + (1 - \alpha)\,\overline{\mathrm{JSD}}_t,
\]
with $\alpha = 0.5$. Higher $D_t$ indicates stronger distributional change between the reference window and evaluation window $t$.

\paragraph{Reliability failure signals.}
DTRC uses the pre-action reliability metrics from window $t$---specifically $\mathrm{ECE}_t$ and $\mathrm{AUC}_t$---to determine whether reliability has degraded sufficiently to warrant intervention. Thresholds $\theta_C$ and $\theta_A$ define alarm levels:
\[
\textsf{calib\_fail}_t = \mathbf{1}\{\mathrm{ECE}_t \ge \theta_C\}, \qquad
\textsf{disc\_fail}_t = \mathbf{1}\{\mathrm{AUC}_t \le \theta_A\}.
\]
These signals ensure that interventions are triggered only when drift coincides with measurable reliability degradation, avoiding unnecessary updates during benign distributional changes.

\paragraph{Decision rule.}
DTRC is parameterized by $(\theta_{d1}, \theta_{d2}, \theta_C, \theta_A)$ and follows a drift-first hierarchical rule. Two drift thresholds partition the drift magnitude into three regimes---low, moderate, and severe---and reliability failure signals determine the intervention type within each regime:
\[
a_t =
\begin{cases}
\text{NoOp}, & \text{if } D_t \le \theta_{d1}, \\[4pt]
\text{Recalibrate}, & \text{if } \theta_{d1} < D_t \le \theta_{d2} \text{ and } \textsf{calib\_fail}_t, \\[4pt]
\text{NoOp}, & \text{if } \theta_{d1} < D_t \le \theta_{d2} \text{ and } \neg\,\textsf{calib\_fail}_t, \\[4pt]
\text{Both}, & \text{if } D_t > \theta_{d2} \text{ and } (\textsf{calib\_fail}_t \lor \textsf{disc\_fail}_t), \\[4pt]
\text{Retrain}, & \text{if } D_t > \theta_{d2} \text{ and } \neg\,\textsf{calib\_fail}_t \land \neg\,\textsf{disc\_fail}_t.
\end{cases}
\]
At the first evaluation window (or if no model exists), the policy performs an initial training step (\text{TrainInit}) using all available historical data up to $t{-}1$.

\begin{algorithm}[H]
\caption{DTRC: Drift-Triggered Reliability Control (drift-first hierarchical)}
\label{alg:dtrc}
\begin{algorithmic}[1]
\Require Pre-action metrics $\mathrm{AUC}_t, \mathrm{ECE}_t$; combined drift signal $D_t$
\Require Thresholds $(\theta_{d1}, \theta_{d2}, \theta_C, \theta_A)$
\If{no model exists}
    \State Train model on available historical data
    \State \Return \textbf{TrainInit}
\EndIf
\State $\textsf{calib\_fail} \leftarrow (\mathrm{ECE}_t \ge \theta_C)$
\State $\textsf{disc\_fail} \leftarrow (\mathrm{AUC}_t \le \theta_A)$
\If{$D_t \le \theta_{d1}$}
    \State \Return \textbf{NoOp}
\ElsIf{$D_t \le \theta_{d2}$} \Comment{Moderate drift}
    \If{$\textsf{calib\_fail}$}
        \State \Return \textbf{Recalibrate}
    \Else
        \State \Return \textbf{NoOp}
    \EndIf
\Else \Comment{Severe drift: $D_t > \theta_{d2}$}
    \If{$\textsf{calib\_fail}$ \textbf{or} $\textsf{disc\_fail}$}
        \State \Return \textbf{Both} (Retrain then Recalibrate)
    \Else
        \State \Return \textbf{Retrain}
    \EndIf
\EndIf
\end{algorithmic}
\end{algorithm}

\subsection{Multi-Objective Reliability Control (MORC)}
\label{sec:morc}

DTRC requires specifying the threshold vector $(\theta_{d1}, \theta_{d2}, \theta_C, \theta_A)$. Rather than selecting these ad hoc, MORC treats thresholded DTRC policies as a parameterized policy class $\Pi_\Theta = \{\pi_\theta : \theta \in \Theta\}$ and searches over $\Theta$ to characterize stability--cost trade-offs systematically.

\paragraph{Threshold construction.}
Drift thresholds $\theta_{d1}$ and $\theta_{d2}$ are drawn from an effective candidate set constructed from midpoints between the sorted unique observed drift values across deployment windows. This produces candidate thresholds that lie at decision boundaries between distinct year partitions, ensuring each candidate yields a potentially different intervention sequence. Reliability thresholds $\theta_C$ and $\theta_A$ are set to fixed quantiles of the empirical ECE and AUC distributions from the baseline (P0) deployment: $\theta_C = q_{0.8}(\mathrm{ECE})$ and $\theta_A = q_{0.2}(\mathrm{AUC})$, corresponding to alarm levels exceeded approximately 20\% of the time under static deployment.

\paragraph{Grid evaluation.}
Each pair $(\theta_{d1}, \theta_{d2})$ with $\theta_{d2} > \theta_{d1}$ defines a valid DTRC policy. Combined with the fixed $(\theta_C, \theta_A)$, the sweep evaluates all such pairs from the candidate set. This yields 69 threshold configurations before deduplication; after removing configurations that produce identical intervention sequences (due to the discrete, annual evaluation structure), 23 distinct operating points remain. Each configuration $\theta$ induces a full deployment trajectory and yields a pair $(V_{L1}(\pi_\theta), C(\pi_\theta))$.

\paragraph{Pareto frontier and knee selection.}
We compute the empirical Pareto frontier of non-dominated configurations in the cost--volatility plane. A configuration $\theta$ is Pareto-dominated if another configuration achieves both lower volatility and lower cost. To instantiate a single operating point for deployment, we select a \emph{knee} configuration by identifying the point on the frontier that minimizes volatility subject to a moderate cost budget ($C \le 15$). This favours configurations that achieve large volatility reductions with limited retraining. In our experiments, the knee configuration corresponds to $\theta_{d1} = 0.029$, $\theta_{d2} = 0.053$, yielding 1 retraining event and 2 recalibrations over 9 annual windows (total cost = 12), compared to P2's 9 retraining events (total cost = 45).

\begin{table}[H]
\centering
\caption{Policy hyperparameters. DTRC thresholds are instantiated using the MORC knee configuration.}
\label{tab:policy_hparams}
\begin{tabular}{ll}
\toprule
Policy & Hyperparameters \\
\midrule
P0 (Static) & No updates; model trained once on pre-2010 data \\
P1 (Periodic recalib.) & Recalibration interval $\Delta t = 1$ year (isotonic, previous-year cohort) \\
P2 (Rolling retrain) & Rolling retraining window $W = 3$ years \\
P3 (DTRC) & $\theta_{d1} = 0.0342,\ \theta_{d2} = 0.0517,\ \theta_C = 0.0937,\ \theta_A = 0.5663$ \\
P4 (MORC) & Grid of $(\theta_{d1}, \theta_{d2})$ pairs (69 candidates, 23 distinct operating points) \\
\bottomrule
\end{tabular}
\end{table}

\section{Experiments}
\label{sec:experiments}

\subsection{Dataset and Prediction Task}
\label{sec:experiments_dataset}

We evaluate the proposed framework on a large-scale, temporally indexed credit-risk dataset of granted loan originations with a binary default outcome \citep{Arroyo2024lending}. The dataset contains 1{,}347{,}681 loans issued between June 2007 and December 2018, with an overall default rate of 19.98\%. We focus on ten covariates that are standard in credit scoring: revenue, debt-to-income ratio (dti\_n), loan amount, normalized FICO score (fico\_n), employment experience category (experience\_c), employment length (emp\_length), purpose, home-ownership (home\_ownership\_n), state (addr\_state), and zip code (zip\_code). The target label is \texttt{Default}.

We construct a deployment horizon covering loans issued from 2010 to 2018. All loans issued before 2010 are used for initial model training and for constructing drift and reliability thresholds.

\begin{table}[H]
\centering
\caption{Dataset summary and temporal evaluation structure.}
\label{tab:dataset_summary}
\begin{tabular}{lr}
\toprule
Statistic & Value \\
\midrule
Total samples & 1{,}347{,}681 \\
Date range & 2007-06 to 2018-12 \\
Number of features & 10 \\
Target variable & Default (binary) \\
Default rate & 19.98\% \\
Evaluation windows ($T$) & 9 (annual, 2010--2018) \\
Training cutoff ($t_0$) & 2009-12-31 \\
\bottomrule
\end{tabular}
\end{table}

\subsection{Temporal Evaluation Protocol}
\label{sec:experiments_protocol}

Evaluation follows a strictly chronological protocol. We set a training cutoff date $t_0 = \text{2009-12-31}$ and define $T = 9$ non-overlapping annual evaluation windows from 2010 through 2018. At each window $t$, the deployed system produces probability predictions for all loans issued in year $t$ and is evaluated on the corresponding labeled outcomes. Any intervention selected after evaluating window $t$ updates the deployed system for subsequent windows ($t+1$ onward), so that decisions never use future information.

For rolling retraining (P2), training data for window $t$ are drawn from the preceding $W$ years only. We fix the retraining horizon to $W = 3$ years, so that, for example, the model evaluated in 2015 is trained on loans issued from 2012 to 2014.

For DTRC and MORC, drift indicators at time $t$ are computed by comparing window $t$ to a reference constructed from recent history (typically the last $W = 3$ years). Drift is thus measured against the most recent regime rather than the entire historical dataset.

\subsection{Base Model and Implementation}
\label{sec:experiments_model}

The base predictor is a probabilistic classifier producing scores in $[0,1]$. We use gradient-boosted decision trees implemented via XGBoost, chosen for their strong performance and robustness on tabular data. Preprocessing, calibration, and evaluation utilities are implemented using \texttt{scikit-learn}. Categorical features are encoded with one-hot encoding, numeric features are median-imputed, and all transformations are wrapped in a single pipeline. Random seeds are fixed to ensure reproducibility across policies.

Key XGBoost hyperparameters (number of trees, depth, learning rate, subsampling, and regularization settings) are held constant across all policies and are reported in Appendix~\ref{sec:appendix_impl}. This isolates the effect of deployment policy design from changes in the underlying model architecture or capacity.

\subsection{Calibration and Updating Mechanisms}
\label{sec:experiments_updates}

Recalibration is implemented via isotonic regression. At evaluation window $t$, when a recalibration action is chosen, we fit an isotonic calibrator on the model’s scores and outcomes from the previous window $t{-}1$ and apply it prospectively to window $t$. This preserves the model’s ranking while adjusting probability estimates to improve calibration.

Retraining refits the underlying XGBoost model on the policy-specified training set. For P2, this is the rolling $W = 3$ year window preceding $t$. For DTRC and MORC, retraining uses a rolling window of comparable length, potentially restricted or extended depending on data availability in earlier years. When both retraining and recalibration are invoked in the same step, the model is first retrained and then recalibrated using labeled data from the most recent window, so that calibration is always applied on top of the current predictor.

\subsection{Metrics}
\label{sec:experiments_metrics}

We evaluate both pointwise predictive performance and the temporal stability of reliability. For each evaluation year $t$ we compute ROC AUC (discrimination), Expected Calibration Error (ECE), and the Brier score. These metrics quantify how well the model ranks borrowers, how well predicted probabilities match observed default frequencies, and the overall quality of probabilistic predictions.

To capture temporal stability, we use the volatility metrics defined in Section~\ref{sec:problem_formulation}. The $L_1$ volatility
\[
V_{L1} = \frac{1}{T-1} \sum_{t=2}^{T} \left( |A_t - A_{t-1}| + |C_t - C_{t-1}| \right)
\]
summarizes the average absolute change in discrimination and calibration between consecutive years. Downside volatility $V^{-}_{L1}$ restricts this sum to steps where reliability degrades (for example, when $A_t < A_{t-1}$ or $C_t > C_{t-1}$). This emphasizes destabilizing movements in reliability that are most relevant for governance and risk management.

\subsection{Drift Indicators and Policy Suite}
\label{sec:experiments_policies}

We evaluate policies P0--P3 and construct P4 by sweeping a family of MORC threshold configurations. The four primary policies are:

\begin{itemize}
  \item P0 (Static): no updates after initial training.
  \item P1 (Periodic recalibration): isotonic recalibration every year using the previous-year cohort.
  \item P2 (Rolling retraining): retraining every year on a rolling $W = 3$-year window.
  \item P3 (DTRC): drift-triggered interventions combining recalibration, retraining, or both.
\end{itemize}

Drift indicators summarize distributional change between the reference window and the current evaluation window. For numeric features, we compute the Kolmogorov--Smirnov (KS) two-sample statistic per feature and take the mean across features. For categorical features, we compute the Jensen--Shannon divergence (JSD) per feature using compressed category histograms (top-$k$ categories plus an \texttt{OTHER} bin) and average across features. The combined drift score
\[
D_t = \alpha \,\overline{\mathrm{KS}}_t + (1 - \alpha)\,\overline{\mathrm{JSD}}_t,
\]
with $\alpha = 0.5$, serves as the input to the DTRC and MORC decision rules.

For P4 (MORC), we treat DTRC as a parameterized policy class and sweep over a grid of drift and reliability thresholds. We first fix calibration and discrimination alarm thresholds $\theta_C$ and $\theta_A$ to empirical quantiles of the static deployment (P0), corresponding to roughly the 80th percentile of ECE and the 20th percentile of AUC. We then generate candidate drift thresholds from midpoints between sorted unique values of $D_t$ across the deployment horizon, add extreme values to allow “always off” and “always on” regimes, and evaluate all ordered pairs $(\theta_{d1}, \theta_{d2})$ with $\theta_{d2} > \theta_{d1}$. This yields 69 candidate configurations before deduplication and 23 distinct operating points in the cost–volatility plane, which we found sufficient to densely cover the relevant part of the frontier. The MORC knee configuration used for P3 is drawn from this frontier.

\subsection{Intervention Cost Accounting}
\label{sec:cost_accounting}

Intervention cost is accounted for using the weighted scheme defined in Section~\ref{sec:problem_cost}. Each retraining event incurs a cost of 5 units, each recalibration incurs a cost of 1 unit, and NoOp actions incur zero cost. When an action combines retraining and recalibration, the costs are additive. The cumulative deployment cost
\[
C(\pi) = \sum_{t=1}^{T} c(a_t)
\]
aggregates these per-window charges. Reporting cost in this way makes the trade-off between stability and intervention intensity explicit: P2, for example, incurs a total cost of 45 over nine years (one retrain per year), while the DTRC knee policy achieves comparable discrimination and lower volatility with only 12 cost units (one retrain and two recalibrations).

\subsection{Uncertainty Estimation}
\label{sec:experiments_bootstrap}

To reflect sampling variability, we quantify uncertainty in volatility metrics using a block bootstrap over yearly windows. We resample the nine annual windows with replacement to form 1{,}000 bootstrap replicates of the reliability trajectory under each policy, recompute $V_{L1}$ and $V^{-}_{L1}$ for each replicate, and report percentile-based 95\% confidence intervals. This procedure preserves the within-year dependence structure while capturing uncertainty in the temporal evolution of reliability.

\FloatBarrier

\section{Results}
\label{sec:results}

We evaluate deployment policies along three complementary axes:
(i) average predictive performance,
(ii) temporal stability of reliability trajectories, and
(iii) cost--volatility trade-offs.

\subsection{Pointwise Performance Across Deployment}

Table~\ref{tab:global_performance} summarizes discrimination, calibration, volatility, and cost aggregated over the nine annual deployment windows (2010--2018).

\begin{table}[H]
\centering
\caption{Global policy comparison aggregated over the full deployment horizon.
Lower is better for ECE, Brier score, volatility ($V_{L1}$, $V^{-}_{L1}$), and cost.
Cost is computed under the weighted scheme (5 per retrain, 1 per recalibration, 0 for NoOp).}
\label{tab:global_performance}
\begin{tabular}{lcccccc}
\toprule
Policy & Mean AUC & Mean ECE & Mean Brier & $V_{L1}$ & $V^{-}_{L1}$ & Cost \\
\midrule
P0 (Static) & 0.608470 & 0.035015 & 0.146761 & 0.032064 & 0.019486 & 0 \\
P1 (Periodic recalibration) & 0.608128 & 0.025923 & 0.144873 & 0.042180 & 0.024017 & 9\\
P2 (Rolling retrain) & 0.661221 & 0.024565 & 0.140004 & 0.030862 & 0.013754 & 45 \\
P3 (DTRC) & 0.641033 & 0.031933 & 0.142921 & 0.028620 & 0.016649 & 12 \\
\bottomrule
\end{tabular}
\end{table}

Rolling retraining (P2) achieves the highest mean discrimination (AUC 0.661) and the lowest mean Brier score, but at the highest cumulative cost (45 under the weighted action model).[file:2] In contrast, Drift-Triggered Reliability Control (P3) incurs a total cost of 12, corresponding to one retraining and two recalibration actions over nine years, and still maintains competitive discrimination (AUC 0.641).[file:2] This represents roughly a \textbf{73\% reduction} in intervention cost relative to P2 while preserving most of the performance gains of continuous retraining.

\begin{figure}[H]
\centering
\includegraphics[width=0.9\linewidth]{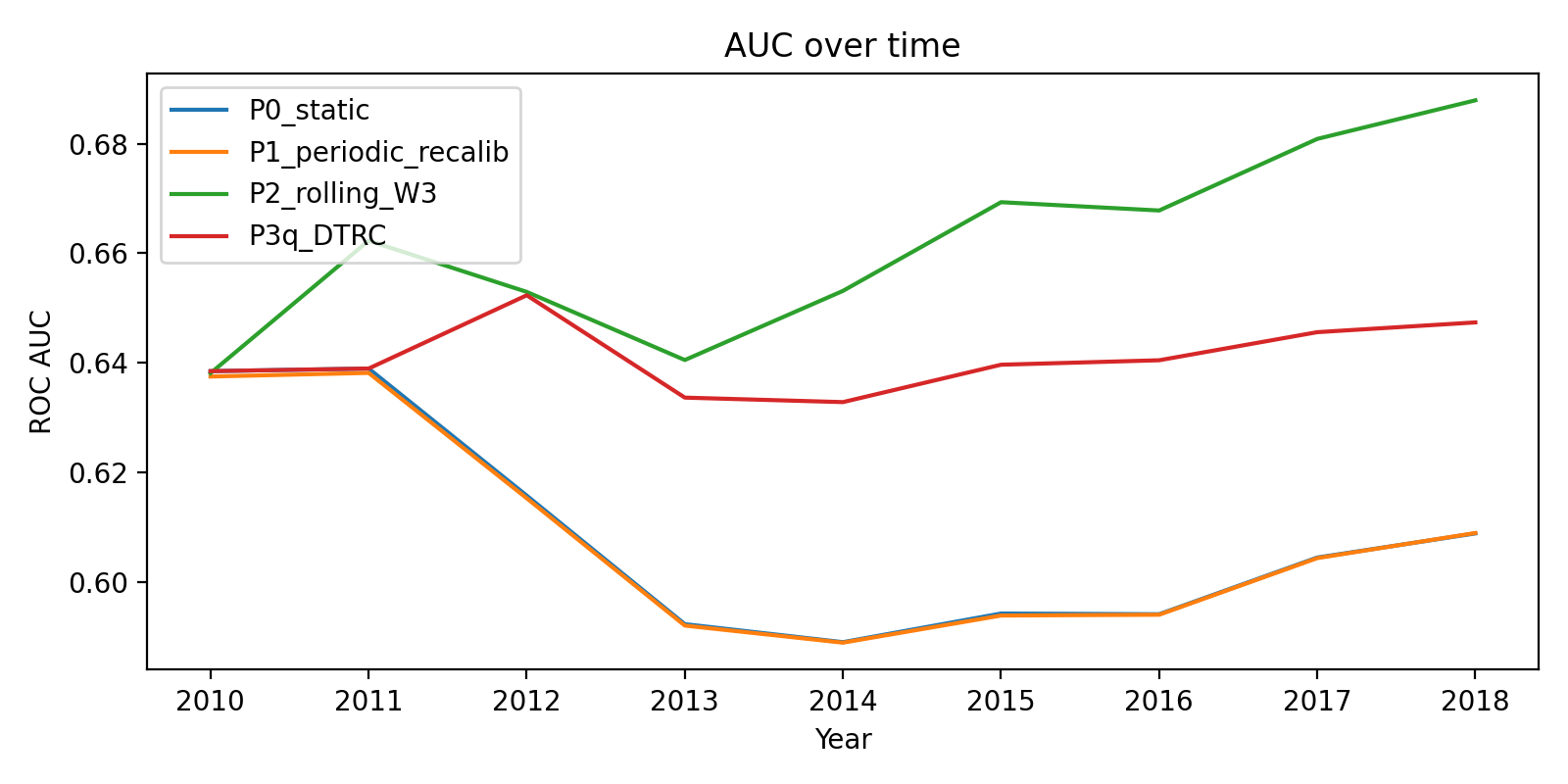}
\caption{ROC AUC across deployment windows. Rolling retraining yields the highest average discrimination but exhibits noticeable fluctuations. DTRC maintains intermediate discrimination while avoiding extreme swings.}
\label{fig:auc_over_time}
\end{figure}

\begin{figure}[H]
\centering
\includegraphics[width=0.9\linewidth]{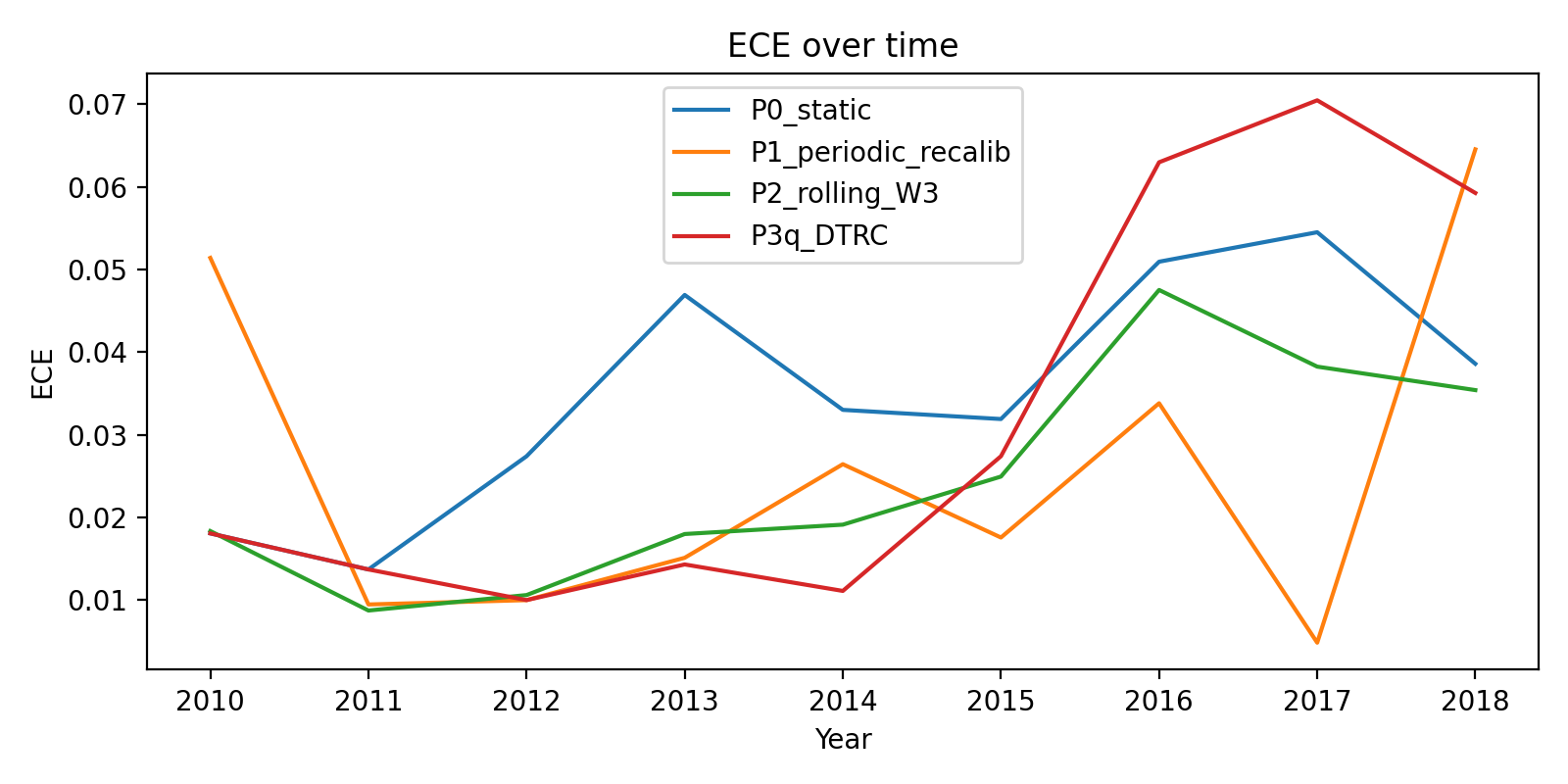}
\caption{Expected Calibration Error (ECE) across deployment windows. Periodic recalibration reduces calibration error on average but introduces oscillatory behaviour across windows; DTRC stabilizes calibration without continuous updates.}
\label{fig:ece_over_time}
\end{figure}

Inspection of Figure~\ref{fig:auc_over_time} shows that static deployment exhibits gradual discrimination erosion after 2012, rolling retraining recovers performance following periods of elevated drift, and DTRC tracks similar recovery phases with substantially fewer updates. The ECE trajectories in Figure~\ref{fig:ece_over_time} further illustrate how different policies shape calibration over time.

\paragraph{Counterintuitive instability of periodic recalibration.}
Periodic recalibration (P1) improves mean ECE relative to static deployment, but this comes with increased temporal variance in calibration error. Notably, P1 exhibits \emph{higher volatility} than static deployment ($V_{L1} = 0.0422$ vs.\ $0.0321$).[file:2] This counterintuitive behaviour reflects oscillations introduced by fitting a fresh isotonic map each year: even when discrimination remains stable, local recalibration adjustments can induce fluctuations in predicted probabilities, thereby increasing trajectory instability. The governance implications of such oscillatory recalibration are discussed further in Section~\ref{sec:discussion}.

\subsection{Reliability Volatility}

Volatility captures instability of the two-dimensional reliability trajectory $(A_t, C_t)$. Static deployment exhibits elevated volatility due to unmitigated temporal drift. Rolling retraining reduces $V_{L1}$ relative to P0, but non-negligible fluctuations persist despite continuous updates.[file:2]

DTRC achieves the lowest $V_{L1}$ among the evaluated policies (0.0286), indicating smoother reliability evolution than even rolling retraining.[file:2] However, in terms of downside volatility $V^{-}_{L1}$, which emphasizes destabilizing reliability degradations, P2 retains a slight advantage (0.0138 vs.\ 0.0166 for P3).[file:2] This suggests that while DTRC smooths overall trajectory variation, continuous retraining may still dampen the most adverse year-to-year changes more aggressively. The trade-off between overall and downside volatility is examined further in Section~\ref{sec:discussion}.

Together, these results show that stability is not monotonically improved by increasing update frequency. Selective intervention can produce lower overall volatility while using substantially fewer retraining events.

\begin{figure}[H]
\centering
\includegraphics[width=0.9\linewidth]{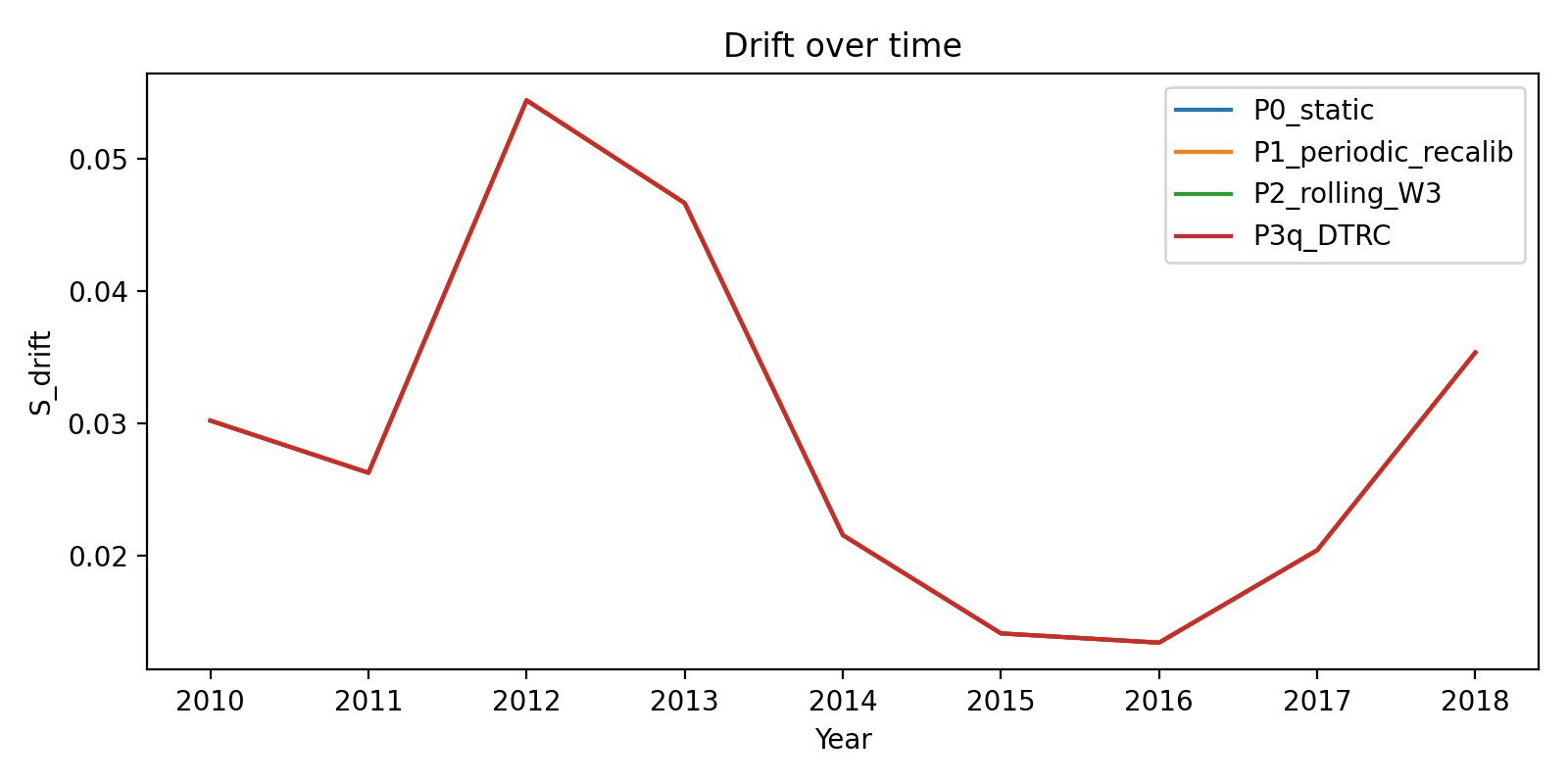}
\caption{Drift signal over time summarizing distributional change between evaluation windows. Periods of elevated drift (e.g., mid-horizon years) coincide with higher reliability volatility under static deployment. DTRC activates selectively during these intervals, smoothing subsequent trajectory evolution.}
\label{fig:drift_over_time}
\end{figure}

\subsection{Cost--Volatility Trade-offs under MORC}

\begin{figure}[H]
\centering
\includegraphics[width=0.9\linewidth]{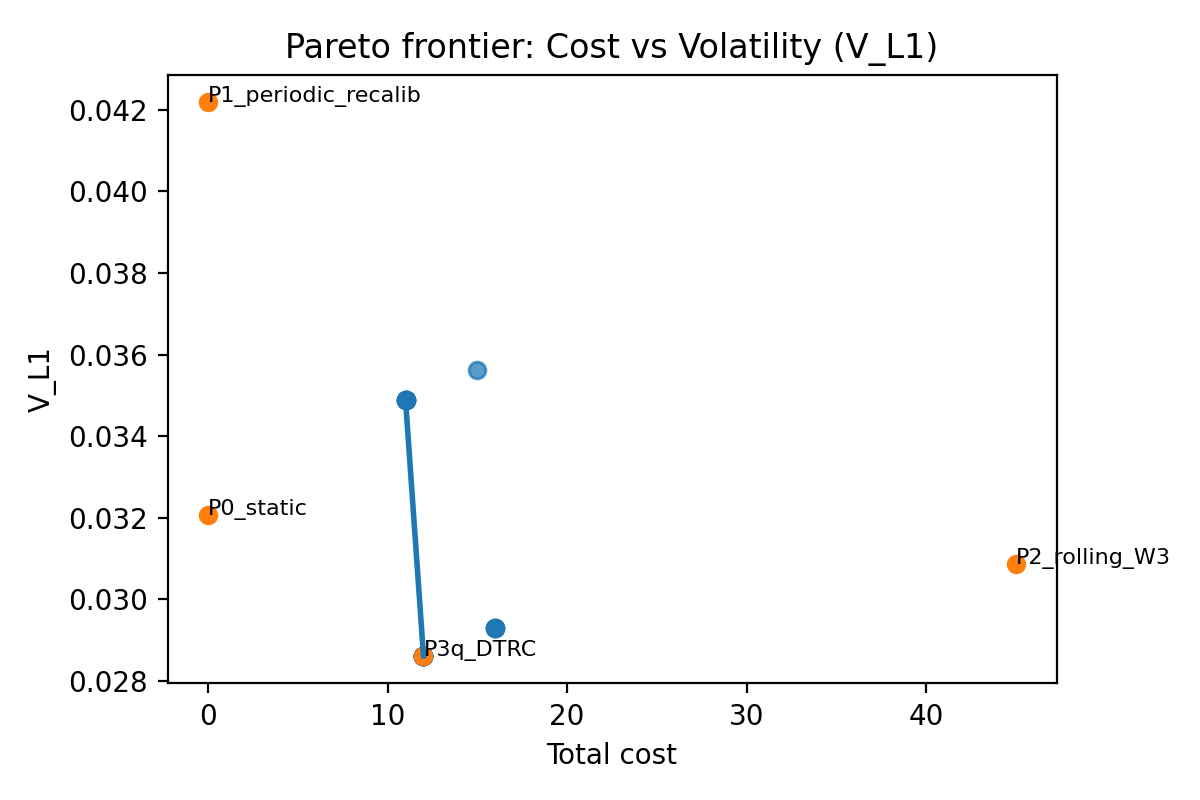}
\caption{Empirical Pareto frontier in cost--volatility space for MORC policies. Points represent distinct threshold configurations; the highlighted point corresponds to the knee configuration used for DTRC.}
\label{fig:p4_pareto_cost_vs_vol}
\end{figure}

To characterize the full stability--cost landscape, we evaluate the MORC policy class over 69 threshold configurations and construct the empirical Pareto frontier (Figure~\ref{fig:p4_pareto_cost_vs_vol}).[file:2]

The frontier exhibits a nonlinear structure: substantial volatility reductions are achievable in the low-cost regime, whereas further gains require disproportionately higher intervention frequency. Table~\ref{tab:pareto_points} reports representative Pareto-optimal policies. The MORC knee (Cost = 12) reduces $V_{L1}$ by around 18\% relative to a nearby minimum-cost configuration (Cost = 11) while adding only a single recalibration action.[file:2]

\begin{table}[H]
\centering
\caption{Representative MORC policies along the cost--volatility frontier.
Cost is computed under the weighted scheme.}
\label{tab:pareto_points}
\begin{tabular}{lccccccc}
\toprule
Label & Mean AUC & Mean ECE & Mean Brier & $V_{L1}$ & $V^{-}_{L1}$ & Cost & Retrains / Recalibs \\
\midrule
P4 (Min-cost) & 0.641010 & 0.026333 & 0.142472 & 0.034894 & 0.016649 & 11 & 1 / 1 \\
MORC-Knee & 0.641033 & 0.031933 & 0.142921 & 0.028620 & 0.016649 & 12 & 1 / 2 \\
\bottomrule
\end{tabular}
\end{table}

The selected DTRC thresholds (Section~\ref{sec:method}) correspond to moderate drift sensitivity: the effective drift thresholds $(\theta_{d1}, \theta_{d2}) \approx (0.028, 0.053)$ indicate that only material distributional shifts trigger intervention, avoiding overreaction to minor fluctuations while still responding to pronounced changes in the data-generating process.[file:2]

\subsection{Bootstrap Robustness}

We assess robustness of volatility comparisons using a block bootstrap over yearly windows (9 temporal blocks from 2010--2018). We resample years with replacement to form 1{,}000 bootstrap replicates for each policy, recompute $V_{L1}$ and $V^{-}_{L1}$, and report percentile-based 95\% confidence intervals.[file:2]

\begin{table}[H]
\centering
\caption{Block bootstrap 95\% confidence intervals for volatility metrics.}
\label{tab:bootstrap_ci}
\begin{tabular}{lcc}
\toprule
Policy & $V_{L1}$ (95\% CI) & $V^{-}_{L1}$ (95\% CI) \\
\midrule
P0 & [0.033, 0.078] & [0.015, 0.041] \\
P1 & [0.036, 0.118] & [0.017, 0.059] \\
P2 & [0.028, 0.066] & [0.013, 0.034] \\
P3 & [0.026, 0.069] & [0.013, 0.035] \\
\bottomrule
\end{tabular}
\end{table}

Confidence intervals partially overlap across policies, reflecting temporal variability across years. However, the lower bound of DTRC’s $V_{L1}$ interval lies below the lower bound for static deployment, and its point estimate remains lower than P2’s despite the much lower cost.[file:2] Given the limited number of yearly blocks, the bootstrap should be interpreted as an exploratory robustness check rather than a formal significance test.

\paragraph{Synthesis.}
Across analyses, continuous retraining maximizes mean discrimination but incurs high operational cost. Periodic recalibration improves mean calibration but can destabilize reliability trajectories. Drift-triggered control achieves lower overall volatility with a large reduction in cumulative intervention cost relative to rolling retraining, while remaining near the Pareto-efficient frontier induced by MORC policies.[file:2] These findings support modeling deployment reliability under temporal shift as a controllable multi-objective system and designing intervention policies explicitly around stability--cost trade-offs.

\section{Discussion}
\label{sec:discussion}

This work reframes deployment under temporal distribution shift as a \emph{controlled} reliability process rather than a sequence of independent evaluations. The empirical findings yield three practical implications for reliability engineering in high-stakes tabular systems.

\paragraph{Average performance is not a sufficient deployment objective.}
Rolling retraining (P2) achieves the strongest mean discrimination, but it does not necessarily yield the most stable reliability trajectory. In operational settings, governance and risk management are driven not only by mean AUC or mean calibration error, but also by the \emph{temporal regularity} of these quantities. Abrupt changes in calibration or discrimination can trigger downstream policy violations, invalidate human decision rules calibrated to prior model behaviour, and undermine trust even when average performance remains high. For instance, a sharp shift in score distributions may require credit policy thresholds or cut-off strategies to be re-approved by risk committees and regulators, adding significant friction even if headline metrics improve.

Prior studies in temporal dataset shift and credit scoring have primarily focused on quantifying performance decay under chronological splits or comparing static retraining heuristics. While such analyses diagnose degradation, they do not explicitly treat deployment as a sequential control problem with stability objectives. Our results suggest that controlling the \emph{trajectory} of reliability is a distinct and practically meaningful objective beyond average predictive performance.

\paragraph{Why periodic recalibration can destabilize reliability.}
A notable result is that periodic isotonic recalibration (P1) improves mean calibration error yet increases volatility relative to static deployment. This suggests that repeatedly fitting flexible, non-parametric calibration maps on small temporal slices may introduce window-to-window oscillation in probability estimates. In practice, this behaviour may be especially problematic when score distributions shift gradually: recalibration corrects local miscalibration but can overreact to short-horizon noise. Practitioners should therefore treat recalibration frequency as a tunable control parameter rather than a fixed annual ritual, adjusting it in light of observed drift, label delays, and governance constraints. This observation highlights that calibration maintenance, when applied mechanically, can itself become a source of instability.

\paragraph{Selective control can dominate frequency-based intervention.}
DTRC (P3) reduces overall reliability volatility relative to rolling retraining while substantially lowering intervention cost. This supports the hypothesis that effective deployment adaptation benefits from \emph{gating} updates using both distributional evidence (drift indicators) and reliability evidence (discrimination and calibration deviations). Importantly, this control framing yields interpretable trade-offs: MORC reveals that large volatility reductions can be obtained in a low-cost regime, whereas further gains require disproportionate retraining intensity. Such trade-off structure is valuable for practitioners because it supports policy selection aligned with operational budgets, model-risk appetites, and review cycles imposed by internal and external oversight.

\paragraph{Downside volatility highlights asymmetric risk.}
Downside volatility captures instability in the direction that increases operational risk (for example, spikes in calibration error). The observation that rolling retraining retains a slight advantage on $V^{-}_{L1}$ while DTRC minimizes overall volatility suggests a natural design space: one may tune control policies depending on whether the deployment context prioritizes overall smoothness or protection against adverse excursions. This motivates future work on explicit asymmetric objectives and risk-sensitive control variants.

Overall, these findings indicate that deployment reliability under temporal shift admits meaningful multi-objective structure and that policy design—not only model choice—determines stability and cost outcomes during real-world operation. Although demonstrated in credit-risk modeling, the framework is directly applicable to other regulated, temporally evolving domains such as electronic health records, insurance underwriting, and public-sector risk assessment, where stability, governance, update cost, label delay, and periodic regulatory audits are central operational concerns.

\section{Limitations}
\label{sec:limitations}

Several limitations contextualize the scope of this study.

\paragraph{Single-domain empirical evaluation.}
Experiments are conducted on a large-scale credit-risk dataset. While this is a representative high-stakes tabular setting, additional evaluation across other temporal domains (for example, claims fraud, churn, or clinical risk scoring) is required to establish how broadly the observed cost--volatility trade-offs generalize.

\paragraph{Policy class and action space.}
We study a restricted family of intervention actions (NoOp, recalibration, retraining, both) and thresholded policies based on drift and reliability indicators. More expressive policy classes could incorporate richer state representations (for example, predictive uncertainty, label-shift estimates, or delayed outcomes), explicit budget constraints, or long-horizon trade-offs. We also do not model fairness or group-wise reliability, so the framework does not yet address trade-offs between stability, cost, and equity. In particular, constrained Markov decision processes or risk-sensitive, fairness-aware control formulations provide natural generalizations of the deployment control perspective adopted here. The MORC sweep should therefore be interpreted as an empirical characterization within a practically motivated design space rather than an exhaustive search over all possible deployment strategies.

\paragraph{Drift signal choice.}
Our drift indicators are based on Kolmogorov--Smirnov statistics for numeric features and Jensen--Shannon divergence for categorical features, aggregated into a single scalar score. These choices emphasize interpretability and ease of implementation, but they do not capture all forms of shift (for example, interaction drift, conditional label shift, or representation shift in learned feature spaces). Alternative drift measures, including kernel-based two-sample tests or model-based shift estimators, may yield different triggering behaviour and merit systematic study.

\paragraph{Bootstrap uncertainty under limited temporal blocks.}
The block bootstrap operates over yearly blocks (nine years). This preserves temporal dependence within each year but yields a limited number of blocks, so confidence intervals should be interpreted as exploratory robustness estimates rather than definitive inferential guarantees. Longer deployment horizons or multiple independent deployment histories would support stronger uncertainty quantification.

\paragraph{Label delay and observation model.}
In practice, default outcomes are observed with a delay after loan origination (for example, at 6- or 12-month horizons), and labels for the most recent windows may be censored or partially observed. Our experiments assume that yearly labels are fully observed at evaluation time, which simplifies the observation model. Extending the framework to explicitly handle label delay and censoring, and to propagate this uncertainty into control decisions, is a natural next step.

\paragraph{Cost model abstraction.}
We model operational cost primarily through retraining and recalibration frequency, reflecting retraining as the dominant engineering and governance burden relative to post-hoc calibration. In practice, costs can also depend on compute budgets, validation workflows, regulatory oversight, and model risk management procedures. Extending the framework to richer, domain-specific cost models is an important direction for deployment realism.

\section{Conclusion}
\label{sec:conclusion}

We introduced a deployment-centric framework that models predictive reliability under temporal distribution shift as a dynamic state and treats temporal stability as a first-class objective alongside average performance. By defining volatility-based measures of discrimination and calibration, and formulating deployment adaptation as a multi-objective control problem over reliability volatility and intervention cost, we move beyond static, pointwise evaluation toward explicit control of reliability trajectories.

Within this framework, we instantiated a family of deployment policies ranging from static and periodic baselines to drift-triggered rules, and used MORC to trace the empirical cost--volatility Pareto frontier. On a temporally indexed credit-risk dataset, drift-triggered control reduced cumulative weighted intervention cost from 45 to 12 units over nine annual deployment windows while maintaining competitive discrimination and achieving smoother reliability trajectories than continuous retraining. These results demonstrate that selective, state-dependent intervention can dominate simple frequency-based update strategies in stability--cost space.

More broadly, our findings show that deployment reliability under temporal shift admits meaningful multi-objective structure and can be shaped by policy design rather than model choice alone. The proposed framework—combining reliability state modeling, volatility objectives, and cost-aware control—offers a template for designing deployment policies in other regulated, temporally evolving domains where stability, governance, update cost, and label delay are central operational concerns. This opens a broader research programme on theoretical guarantees, adaptive intervention policies, and fairness-aware reliability control under temporal shift.

\section{appendix}

\subsection{Implementation Details}
\label{sec:appendix_impl}

\paragraph{Software stack and environment.}
All experiments are implemented in Python using \texttt{scikit-learn} for preprocessing, calibration, and evaluation, and \texttt{XGBoost} for gradient-boosted trees. Data manipulation uses \texttt{pandas} and \texttt{numpy}. Experiments were run on a Linux server with 64 GB RAM; no GPU acceleration is required for the reported results.

\paragraph{Preprocessing and feature handling.}
We use ten covariates from the credit-risk dataset (Section~\ref{sec:experiments_dataset}). Categorical features (\texttt{experience\_c}, \texttt{purpose}, \texttt{home\_ownership\_n}, \texttt{addr\_state}, \texttt{zip\_code}) are one-hot encoded. Numeric features (revenue, dti\_n, loan amount, fico\_n, emp\_length) are median-imputed. All preprocessing is encapsulated in a single pipeline with the XGBoost classifier, ensuring that the same transformations are applied consistently across policies and time windows.

\paragraph{Base model hyperparameters.}
The same XGBoost configuration is used for all policies to isolate the effect of deployment decisions. Table~\ref{tab:xgb_hparams} summarizes the hyperparameters.

\begin{table}[H]
\centering
\caption{XGBoost hyperparameters used in all experiments.}
\label{tab:xgb_hparams}
\begin{tabular}{ll}
\toprule
Hyperparameter & Value \\
\midrule
\texttt{n\_estimators} & 200 \\
\texttt{max\_depth} & 4 \\
\texttt{learning\_rate} & 0.05 \\
\texttt{subsample} & 0.9 \\
\texttt{colsample\_bytree} & 0.9 \\
\texttt{min\_child\_weight} & 5 \\
\texttt{reg\_lambda} & 1.0 \\
\texttt{objective} & \texttt{binary:logistic} \\
\texttt{eval\_metric} & \texttt{logloss} \\
\texttt{tree\_method} & \texttt{hist} \\
\texttt{max\_bin} & 256 \\
\texttt{random\_state} & fixed seed \\
\bottomrule
\end{tabular}
\end{table}

\paragraph{Temporal splits and evaluation windows.}
We define a training cutoff at 2009-12-31 and construct $T = 9$ non-overlapping annual evaluation windows from 2010 to 2018. For each year $t$, the deployed model produces probability predictions for all loans issued in that year and is evaluated on the corresponding outcomes. Under rolling retraining (P2), the training set for year $t$ consists of loans from the preceding $W = 3$ years when available.

\paragraph{Drift indicators.}
Drift between a reference period and year $t$ is computed using:
(i) the mean Kolmogorov--Smirnov statistic across numeric features, and
(ii) the mean Jensen--Shannon divergence across categorical features, with category histograms compressed to the top-$k$ categories ($k=50$) plus an \texttt{OTHER} bin.
The combined drift score $D_t$ is a convex combination of these two components with $\alpha = 0.5$.

\paragraph{Calibration implementation.}
Isotonic regression is applied through \texttt{sklearn.isotonic.IsotonicRegression} on predicted scores and observed outcomes from the previous year $t{-}1$. When a recalibration action is selected, the fitted calibrator is wrapped around the current model and used for all subsequent predictions until the next calibration or retraining event.

\paragraph{Reliability metrics and volatility computation.}
For each year $t$, ROC AUC, ECE, and Brier score are computed using \texttt{scikit-learn} utilities with 15 equal-width probability bins for ECE. Volatility metrics $V_{L1}$ and $V^{-}_{L1}$ are computed from the annual sequences $\{A_t\}_{t=1}^T$ and $\{C_t\}_{t=1}^T$ as defined in Section~\ref{sec:experiments_metrics}, using simple vectorized operations over the per-year metrics.

\paragraph{Cost model and logging.}
The weighted cost model assigns cost 0 to NoOp, 1 to recalibration, 5 to retraining, and 6 to combined retrain+recalibrate. At each year $t$, we log the chosen action, whether retraining or recalibration occurred, the per-step cost, the drift score $D_t$, and pre- and post-intervention reliability metrics. Cumulative cost is the sum of per-year costs over the nine-year horizon.

\paragraph{Bootstrap procedure.}
For the block bootstrap, we treat each calendar year as a block and resample the nine years with replacement to construct 1{,}000 bootstrap trajectories per policy. Volatility metrics are recomputed on each resampled trajectory, and percentile-based 95\% confidence intervals are derived from the empirical distributions.

\bibliographystyle{apalike}
\bibliography{references}
\end{document}